\documentclass[11pt,letterpaper]{article}
\usepackage{cogsys}
\usepackage[T1]{fontenc}
\usepackage{times}
\usepackage[pdftex]{graphicx} 

\usepackage{natbib}
\usepackage{graphicx}
\usepackage{algorithm}
\usepackage{algorithmic}
\usepackage{array}
\usepackage{pgfplots}
\usepackage{pgfplots, pgfplotstable}
\usepackage{amsmath}
\usepackage{verbatim}

\ShortHeadings{Toward Givenness Hierarchy Theoretic Natural Language Generation}
            {P.\ Pal and T.\ Williams}

\begin{document} 

\title{Toward Givenness Hierarchy Theoretic Natural Language Generation}
 
\author{Poulomi Pal}{poulomipal@mines.edu}
\address{MIRRORLab Colorado School of Mines, 1600 Illinois Street, Golden, CO 80401 USA}
\author{Tom Williams}{twilliams@mines.edu}
\address{MIRRORLab Colorado School of Mines, 1600 Illinois Street, Golden, CO 80401 USA}
\vskip 0.2in
 
\begin{abstract}Language-capable interactive robots participating in dialogues with human interlocutors must be able to naturally and efficiently communicate about the entities in their environment. A key aspect of such communication is the use of anaphoric language.
The linguistic theory of the  \textit{Givenness Hierarchy} (GH) suggests that humans use anaphora
based on
the \textit{cognitive statuses} their referents have in the minds of their interlocutors. 
In previous work, researchers presented GH-theoretic approaches to robot anaphora understanding.
In this paper we describe how the GH might need to be used quite differently to facilitate robot anaphora generation.
\end{abstract}

\section{Introduction} 
 
For language-enabled collaborative robots to be effectively introduced into human society, they must be able to communicate about the objects, locations, and people in human environments 
and must be able to understand and use pronominal forms like \textit{it}, \textit{this}, and \textit{that} just like humans. According to the theory of the \textit{Givenness Hierarchy} (GH) \citep{gundel1993cognitive}, when humans use anaphora rather than definite description, they signal their subconscious assumptions about the \textit{cognitive status} their target referent has in the mind of their interlocutor. 
Thus, for robots to understand and generate human-like language they must be able to model the cognitive status of their referents.

Previously,  \citet{williams2018reference} (see also~\cite{williams2016situated}) presented the first full computational implementation of the GH for the purpose of robotic natural language understanding (NLU), using a set of hand-crafted rules informed by the GH literature.  
Specifically, while \citet{williams2018reference} demonstrated that for anaphora understanding what is needed is only a set of searchable GH-theoretic data structures, as we will show in this paper, efficient anaphora generation requires these data structures to be augmented with a model that directly maps (some limited set of) entities to their presumed cognitive status.

In previous work \citep{pal2020cogsci}, we proposed a statistical per-entity model of cognitive status in which a \textit{Cognitive Status Engine} comprised of per-entity \textit{Cognitive Status Filters} is used to maintain a distribution over cognitive statuses for each entity the robot believes to be at least familiar to all parties within the conversation. In this paper, we describe how this model can be used in conjunction with data structure oriented methods to facilitate efficient and effective algorithms for deciding when and how to use anaphora. 
%


\section{Related Work}\label{sec:relatedwork}
\subsection{Cognitive Status}

The Givenness Hierarchy, originally presented by \citet{gundel1993cognitive}, is comprised of six hierarchically nested tiers of cognitive statuses: \textit{in focus, activated, familiar, uniquely identifiable, referential, and type identifiable}, associated with each of which is a set of referring (or pronominal) forms that can be used when referring to an entity with that status \citep{gundel2006coding,hedberg2013applying}. 
The GH coding protocol, presented by \citet{gundel2006coding}, provides guidelines as to what features of linguistic and environmental context can be used to determine the cognitive status of a given entity. 
Since the GH has been shown to be well validated across a wide variety of languages beyond English \citep{gundel2010testing}, many researchers such as \citet{kehler2000cognitive, chai2004probabilistic,chai2006cognitive,williams2016situated} and~\citet{williams2018reference} have sought to computationally implement it for the purposes of reference resolution. These approaches use the GH to (1) justify sets of data structures that hold representations of candidate referents with different presumed cognitive statuses, and (2) determine how to use those data structures when resolving referring expressions. 

While this data structure-oriented approach facilitates the efficiency of NLU by providing restricted subsets of candidates to be considered, during referring expression \textit{generation}, the speaker already knows what entity they wish to refer to, and needs to determine what cognitive status that entity likely has in order to determine how to refer to it. 
Thus, what is needed, during natural language generation (NLG) (in \textit{addition} to these data structures, as we will later describe) is a means of quickly determining not what entities have a given cognitive status, but what cognitive status is most likely for a given entity. 
Accordingly, the first contribution of this paper is to fill this gap in the literature by proposing a technique for jointly maintaining GH-theoretic cognitive status buffers needed for efficient language understanding and per-entity cognitive status models needed for NLG.

\subsection{Referring Expression Generation}
A key task in NLG is referring expression generation (REG): deciding how to refer to an entity so as to disambiguate it from a contrast set. 
Recent work in robot dialogue systems has sought to extend classic REG algorithms to handle the knowledge constraints imposed in realistic robotic domains and by realistic robotic architectures~\citep{williams2017referring}. 
Central to these algorithms is identification of the set of distractors that must be eliminated by a generated referring expression. Critically, while these algorithms may naively use as a set of distractors all objects that the speaker knows of, this is typically unneccessary, due to the way that attentional focus, and discourse context often work together to constrain the set of reasonable referents~\citep{grosz1986attention}.

We argue that the GH stands to play a critical role in mediating the interaction between low-level dialogue features and the set of distractors that must be ruled out by REs.
For example, when a speaker uses ``this $NP$", it indicates that the speaker assumes their target referent to be ``Activated" to the listener; a GH-theoretic inspired REG algorithm could thus determine that only entities assumed to be at least Activated must be ruled out in the generated $NP$. However, to the best of our knowledge, the GH has not previously been used to inform REG algorithms in any way.
Accordingly, the second contribution of this paper is to fill this gap in the literature by proposing a technique for using the GH to inform the set of distractors to be eliminated by REG algorithms.

\subsection{Anaphora Generation} 

While much research has been performed on referring expression generation, relatively little attention has been paid to anaphora generation~\cite{gatt2018survey}. Previous work on anaphora generation has explored factors such as discourse structure, coherence, and salience \citep{poesio2004centering,mccoy1999generating,kibble2004optimizing,ge1998statistical,callaway2002pronominalization,mccoy1999generating}, but these works directly associate these low-level features with pronoun use, rather than using cognitive status as a mediating factor as suggested by the linguistics literature. 



The GH's claim that choice of referring form depends on presumed cognitive status~\cite{rosa2011role} suggests that the GH should be a powerful tool for deciding when and how to use anaphoric expressions. 
Naively, one might assume the straightforward use the fact that each tier of the GH is associated with a different set of referring forms that can be used to refer to objects assumed to have that status (or higher). However, because progressively lower tiers become associated with progressively greater numbers of possible referring forms, and because forms like "it", "this", and "that" can be ambiguous, these tier-form associations are not sufficient on their own for language generation. Indeed, although the knowledge of cognitive status affects language understanding and generation, no significant linguistic work exists that identifies the mechanisms by which it is actually \textit{used} for selection of referring forms~\cite{arnold2016explicit}.
Accordingly, the third contribution of this paper is to fill this gap in the literature by proposing a GH-theoretic model of NLG that leverages the GH-theoretic cognitive status modeling and REG capabilities described above.

\section{Cognitive Status Estimation and Memory Modeling} \label{sec:status}
As described in~\cite{pal2020cogsci}, we formulate cognitive status modeling as a Bayesian filtering problem. Let a dialogue consist of a set of utterances $U_0,...,U_n$. For object $o$, let $S_{o}^{t} \in \{I,A,F\}$ denote the cognitive status of $o$ at a particular timestep $t$ after utterance $U_t$ (either In Focus, Activated or Familiar), and let $L_{o}^{t} \in \{N,M,T\}$ denote the linguistic status of $o$ in utterance $U_{t}$ (either not mentioned in the utterance, mentioned in the utterance in a non-topic role, or mentioned in the utterance in a topic role). Using this formalism, our goal is to recursively estimate, for a given object, the probability distribution over cognitive statuses for object $o$ at time $t$:
\begin{equation}
    p(S_o^t) = p(S_o^{t-1})p(L_o^{t})p(S_o^t \mid S_o^{t-1}, L_o^{t}) \label{eq:1}
\end{equation}

\begin{algorithm} \caption{GetStatus(O)} \label{alg:getstatus}
\begin{algorithmic}[1]
\STATE {CSE: Cognitive Status Engine comprised of Cognitive Status Filters $\{CSF_0\dots CSF_n\}$}
\STATE\textbf{if} {$\exists (CSF\in CSE \mid CSF.entity = O)$} \textbf{return} $\underset{S}{argmax} P(S|O)$
\STATE \textbf{else return} $UID$
\STATE \textbf{endif}
\end{algorithmic}
\end{algorithm}

We define a Bayesian filter of this form as a \mbox{\textit{Cognitive Status Filter}} (CSF) for a given object $o$. Given a set of known objects, $O = \{o_1,...,o_n\}$, our goal is then to estimate this distribution for each $o \in O$ at each time step~\citep{pal2020cogsci}. 
We maintain distributions over cognitive statuses for each object presumed to be Familiar or higher through a Cognitive Status Engine (CSE) comprised of CSFs for each such object
(if an entity is not tracked by a CSF, it is assumed to be at most Uniquely Identifiable (UID)). 
We also propose to maintain a set of data structures associated with each tier of cognitive status, 
each of which contains pointers to the objects assumed to have (at most) that status. Whenever the most probable cognitive status for an object changes, this reference is updated. 
The GetStatus(\textit{O}) function (Algorithm \ref{alg:getstatus}) takes an object \textit{O} and returns its most likely cognitive status. If no CSF exists for $O$ within the CSE, ``UID'' is returned; otherwise the most probable cognitive status for $O$ (as determined by the distribution maintained by $O$'s CSF) is returned.

\begin{algorithm} \caption{Describe (O)} \label{alg:describe}
\begin{algorithmic}[1]
\STATE S = GetStatus(O)\label{line:getstatus}
\STATE D = GetDistractors(O,S)\label{line:getdistractors}
\IF {S == ``UID"}\label{line:UID1}
    \RETURN ``the" + REG(O,D)
\ELSIF {S == ``FAM"}
    \RETURN ``that" + REG(O,D)\label{line:FAM2}
\ELSIF {S == ``ACT"}
    \IF {D == $\emptyset$}\label{line:DNZ1}
        \STATE \textbf{if} {dist(O) == close} \textbf{return} ``this'' \textbf{else return} ``that'' 
    \ELSE 
        \RETURN DREG(O,D)
    \ENDIF
\ELSIF{S == ``IF"}
    \STATE \textbf{if} {D == $\emptyset$}\label{line:DNZ2} \textbf{return}``it" \textbf{else return} DREG(O,D)
\ENDIF
\end{algorithmic}
\end{algorithm}

\section{Algorithmic Approach}\label{sec:approach}
We now propose the \textit{Describe} algorithm (Alg.~\ref{alg:describe}).
Our ultimate goal is to decide how to refer to a target $O$ based on its cognitive status (Line~\ref{line:getstatus}). 
We do so using the concept of a \textit{Cognitive Status Engine} (Sec.~\ref{sec:status}, Alg.~\ref{alg:getstatus}). 
Once we have determined the cognitive status of $O$, we use it to decide how to refer to $O$. Each tier of cognitive status is associated with a set of referring forms. One might thus naively assume that to decide how to refer to $O$, one could simply use a referring form associated with $O$'s presumed cognitive status.
While we may do so when $O$ is at most Familiar or Uniquely Identifiable (Lines~\ref{line:UID1}-\ref{line:FAM2}), when $O$ is In Focus or Activated, this is infeasible for two reasons. 
First, there may be multiple candidate referents that could plausibly be referred to using that referring form (e.g., when a form like ``this'' is used to cue an activated object, and more than one activated object exists). 
Accordingly, when this type of ambiguity is identified (i.e., when the set of status-sensitive distractors for an object are non-empty 
)\footnote{To construct a status-sensitive distractor set, we must identify all objects with at least the same status as our target. The cognitive status of the target can be quickly retrieved from the CSE (Section \ref{sec:status}). The distractors to be ruled out are then the union of the data structures associated with the target's current most probable cognitive status and higher.} (Lines~\ref{line:DNZ1} and~\ref{line:DNZ2}), it may be avoided by selecting a referring form \textit{that involves a definite description} at either the same level or a lower level of cognitive status, e.g. selecting ``this $NP$'' rather than "it". 
Second, some referring forms (e.g., "this" and "that") are only appropriate for objects that are physically (or cognitively\footnote{An entity may be ``cognitively close'' if it is reachable within a certain number of edges from topics currently in focus or activated within a semantic memory network, or if it is sufficiently recent within episodic memory.}) close.
When this type of conflict is identified\footnote{
I.e., when the weighted sum ($dist(O)$) over physical and episodic distance measures to object $O$ (either globally or with respect to some contrast set) crosses some threshold}, it may be avoided by selecting a referring form that \textit{does not violate such constraints} at either the same level or a lower level of cognitive status, e.g., selecting "that $N$" rather than "this $N$", as shown by the distance-sensitivity displayed both in Alg.~\ref{alg:describe} and Alg.~\ref{alg:dreg}.

Once we have determined what referring form to use, if that form includes a constituent $NP$, we must generate that referring expression using an REG algorithm such as the Incremental Algorithm \citep{dale1995computational} or DIST-PIA \citep{williams2017referring}, which require (1) a target referent, and (2) a set of distractors that must be ruled out. Critically, because we are using a cognitive status theoretic approach, once we decide what referring form to use for an object, we can intelligently identify a reduced set of distractors to rule out, as we can expect that the listener will not even consider objects of lower cognitive status than that cued by the chosen referring form (Alg.~\ref{alg:dreg}).

\begin{algorithm} \caption{DREG(O,D)} \label{alg:dreg}
\begin{algorithmic}[1]
\STATE \textbf{if} {dist(O)== close} \textbf{return} ``this" + $REG(O, D_{this})$
\STATE \textbf{else if} {dist(O) == far} \textbf{return} ``that" + $REG(O, D_{that})$
\STATE \textbf{else return} ``the" + $REG(O,D)$
\STATE \textbf{endif}
\end{algorithmic}
\end{algorithm}

\section{Conclusion}\label{sec:conclusion}
We have presented techniques for: (1) jointly maintaining GH-theoretic inspired data structures required for effective NLU and the statistical per-entity cognitive status models for NLG, (2) using the GH to identify distractors to be ruled out during REG, and (3) GH-theoretic NLG.
%
In future work we plan to integrate and evaluate all described elements that are not yet implemented within our robotic architecture, to enable effective and efficient anaphora generation in realistic HRI scenarios. We will evaluate our algorithm using a human subject experiment where we will judge human preference for referring forms generated by the algorithm (and presented through a robot) and assess the accuracy and efficiency of those referring forms.



\vspace{-0.25in}

{\parindent -10pt\leftskip 10pt\noindent
\bibliographystyle{cogsysapa}
\bibliography{references}
}


\end{document}